# StyleStegan: Leak-free Style Transfer Based on Feature Steganography

Xiujian Liang 1, Bingshan Liu 2, Qichao Ying 3, Zhenxing Qian 4* and Xinpeng Zhang 5
1 Fudan University, China
E-mail: xiujianliang@ieee.org
2 Engineering Research Center of Digital Forensics, China
E-mail: 202183290228@nuist.edu.cn
3 Fudan University, China
E-mail: 20110240050@fudan.edu.cn
4* Fudan University, China
E-mail: zxqian@fudan.edu.cn
5 Fudan University, China
E-mail: zhangxinpeng@fudan.edu.cn

*Abstract* — In modern social networks, existing style transfer methods suffer from a serious content leakage issue, which hampers the ability to achieve serial and reversible stylization, thereby hindering the further propagation of stylized images in social networks. To address this problem, we propose a leak-free style transfer method based on feature steganography. Our method consists of two main components: a style transfer method that accomplishes artistic stylization on the original image and an image steganography method that embeds content feature secrets on the stylized image. The main contributions of our work are as follows: 1) We identify and explain the phenomenon of content leakage and its underlying causes, which arise from content inconsistencies between the original image and its subsequent stylized image. 2) We design a neural flow model for achieving loss-free and biased-free style transfer. 3) We introduce steganography to hide content feature information on the stylized image and control the subsequent usage rights. 4) We conduct comprehensive experimental validation using publicly available datasets MS-COCO and Wikiart. The results demonstrate that StyleStegan successfully mitigates the content leakage issue in serial and reversible style transfer tasks. The SSIM performance metrics for these tasks are 14.98% and 7.28% higher, respectively, compared to a suboptimal baseline model.

## I. Introduction

With the rapid development of Internet social networks and AI image generation technology, multimedia communication through the use of digital images has become a prevalent method for information exchange. Style transfer techniques [1] have gained significant popularity in image processing software such as TikTok and Photoshop, allowing users to transform ordinary digital images into captivating artistic styles. However, these existing style transfer methods inadvertently introduce modifications to the subtle textures and content information within the images. Consequently, when performing serial style transfers with a stylized image, severe corruption and artifacts occur within the content features, giving rise to the content leakage phenomenon [2], as illustrated in Figure 1. The original content becomes hardly recognizable to the naked eye after multiple style transfers. Furthermore, current de-styling methods face challenges in accurately extracting the original content information directly from the stylized image, often resulting in circular inconsistencies during the image reconstruction process [3].

The presence of content leakage hinders the achievement of high-quality serial style transfer and reverse style transfer, thereby impeding the widespread dissemination of stylized images on social networks. Furthermore, the increasing attention from social users to privacy security and copyright protection issues arising from the dissemination of stylized images in social networks highlights the need for an effective solution to control subsequent stylized permissions. This study aims to address two key problems:

1) How to achieve leak-free style transfer in social networks?
2) How to effectively control the subsequent usage rights of stylized original content?

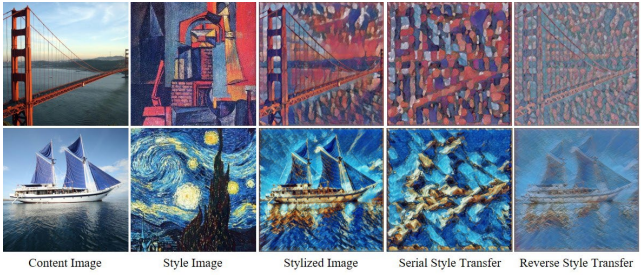

Fig. 1    Content leakage phenomenon in style transfer.

To address the first problem, we have designed a reversible neural flow network as a fundamental solution for achieving leak-free style transfer. This network eliminates content reconstruction loss during the encoding-decoding process and mitigates feature offset errors in the style transformation process. To tackle the second problem, we have introduced image steganography as a means to embed content features into stylized images. This approach enables precise control over subsequent stylized content usage rights. Only social users with specific extraction methods can extract the embedded content features, allowing them to perform high-quality stylization while preserving the original content. By integrating the aforementioned methods, we have successfully resolved the issues of content leakage and subsequent control of stylized images in existing social networks. This achievement not only allows for the more seamless and controlled dissemination of stylized images within social networks, ensuring the preservation of their artistic effects and maximizing their impact during dissemination, but also hold significant practical implications for the diverse transmission of world art.

## II.    RELATED WORK

The successful application of deep learning in computer vision tasks has opened up new possibilities for image style transfer, as initially proposed by [4] using the concept of image iteration. This approach utilized pre-trained convolutional neural networks to extract content and style features and encode them into semantic representations. Building upon this foundation, [1] further advanced the previous method [4] by incorporating additional control over spatial location, color information, and cross-space scales, resulting in significant improvements in the quality of stylized images. However, it is important to note that image iterative methods suffer from slow stylization due to the iterative pixel-by-pixel optimization required for matching style features between the output and input images.

To accelerate the style transfer process, [5] introduced a feedforward model approach aimed at achieving real-time style transfer by training feedforward networks to approximate the iterative optimization process. However, this approach prioritized transfer speed at the expense of image quality degradation. In order to address the trade-off between transfer speed and image quality, [6] made significant improvements to the network proposed in [5] by replacing the batch normalization layer with an instance normalization layer. Additionally, [7] introduced conditional instance normalization to enable real-time style transfer across multiple styles learned during training. Despite these advancements, both image iteration-based and feedforward model-based style transfer methods faced a common challenge in practical applications: they were limited to specific styles and struggled to generalize to arbitrary stylizations. To overcome these limitations, the Adaptive Instance Normalization (AdaIN) method [8] was developed. [8] investigated the impact of statistical properties of stylized image features on the stylization process and proposed an affine transformation based on the mean and standard deviation of stylized images. AdaIN had a significant influence on subsequent research in style transfer, leading to the proposal of various general style transfer methods [9-16]. As a result, style transfer entered the era of arbitrary style transfer.

In practical applications on social networks, stylized images have the potential to alter the original content information. This leads to a significant issue of content leakage when social users perform serial and reverse stylization on these stylized images. While [17] focused on de-styling, their work only addressed face de-styling. To tackle this problem, [18] introduced STNet, a steganographic network based on style transfer, which embeds secret information within the style features of the stylized image. Building upon this, [19] proposed PSTNet, which enhances embedding capacity and steganalysis resistance during the stylization process. Motivated by advancements in steganography [32-34], Chen et al. [20] attempted to embed content image features on stylized images, while [21] explored hiding content image features among layered stylized features. However, due to the inherent feature loss and offset caused by encoding-decoding operations in style transfer, the content features still experience changes in texture details and spatial locations. Existing image steganography-based style transfer methods only address the loss of content from the stylized images generated after transfer, failing to provide a fundamental solution to the content leakage problem that occurs during the transfer process.

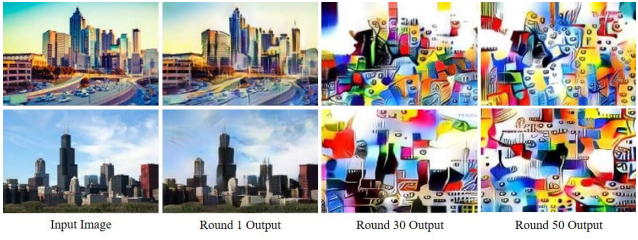

Fig. 2    Test results of iterative inference with the AdaIN

## III. PRE-ANALYSIS

### A. Reasons

To address the content leakage problem, we first investigate and summarize the causes, which can be categorized as follows:

1. Lossy image reconstruction

The model architecture commonly adopted by existing methods for style transfer consists of encoder-decoder pairs. This architecture is based on the intuitive explanation that the decoders of existing style transfer methods cannot achieve lossless image reconstruction of the input content images. To illustrate this point, three methods [22-24] use VGG19 as an encoder and train a decoder to invert the features of VGG19 into the image space. However, despite using image reconstruction loss or content loss functions to train the decoder, it was found in [23] that the feature inversion is imperfect due to the loss of spatial information during the pooling operation in the encoder. Consequently, the accumulated errors in image reconstruction can gradually interfere with the content details and result in content leakage.

2. Biased decoding training

In addition to the image reconstruction loss, which can explain part of the content leakage phenomenon, the biased decoder training scheme is another major contributing factor. In this case, we will use the training scheme of AdaIN as an example to explain how its loss function is configured in a way that leads to content leakage. Specifically, AdaIN trains the decoder using a weighted combination of content loss and style loss, where

$$L_c = \|F(G(t)) - t\|_2 \quad (1)$$

$$L_s = \sum_{i=1}^{L} \|\mu(\varphi_i(G(t))) - \mu(\varphi_i(s))\|_2 + \sum_{i=1}^{L} \|\sigma(\varphi_i(G(t))) - \sigma(\varphi_i(s))\|_2 \quad (2)$$

Figure 2 illustrates the process of style transfer using the AdaIN self-encoder. Without any style image for transfer guidance, a content image is given as input. The self-encoder, which is already pre-trained, performs 50 rounds of style transfer iterations. Each round involves using the output of the previous round's decoder as the input for the next round's encoder, leading to iterative image reconstruction. As the number of inference rounds increases, the generated results start exhibiting strange artistic patterns. This phenomenon suggests that during training, AdaIN's encoder is biased towards style rendering. Moreover, during inference, it gradually becomes more biased towards style.

3. Biased transfer module

In addition to the image reconstruction loss and decoding training bias, the transfer principle of some style transfer modules itself introduces a bias for the content features. Let's take the style decorator module in Avatar-Net as an example. The style decorator, inspired by deep image analogy [25], consists of two main steps. In the first step, the algorithm identifies a corresponding feature block in based on the content similarity between two feature blocks from . Then, in the second step, it finds a corresponding feature block in by replacing the corresponding feature block in . This replacement operation creates a feature block that cannot be recovered from , resulting in a bias towards stylistic features and content leakage.

### B. Remedies

Based on the above analysis, realizing loss-free and bias-free style transfer is a necessary prerequisite to solve the content leakage problem. To achieve loss-free feature extraction, we propose a feature extraction network based on the neural flow model. This network maps the content features and style features of an image into the latent space in a loss-free manner, which is then inputted to the style transfer module. The specific details of the network will be introduced in Section IV. In addition, to achieve bias-free style transfer, we introduce the AdaIN-based style transfer module. This module ensures separable and bias-free image content and style features. The reasons for introducing this module will be described below.

The mechanism of the popular generic style transfer methods at this stage can be seen as a natural evolution of the bilinear model [26]. The model separates image features into content factors and style factors, and then performs style transfer by replacing the style factors in the content image with the style factors in the target image.

**Definition 1** *Suppose there is a bilinear style transfer module* $f_{cs} = C(f_c)S(f_s)$, *where* $C$, $S$ *denotes the content factor and style factor in the bilinear model, respectively. If* $C(f_{cs}) = C(f_c)$ *and* $S(f_{cs}) = S(f_s)$, *then* $f_{cs}$ *is an unbiased style transfer module.*

**Theorem 1** *The adaptive instance normalization in AdaIN is an unbiased style transfer module.*

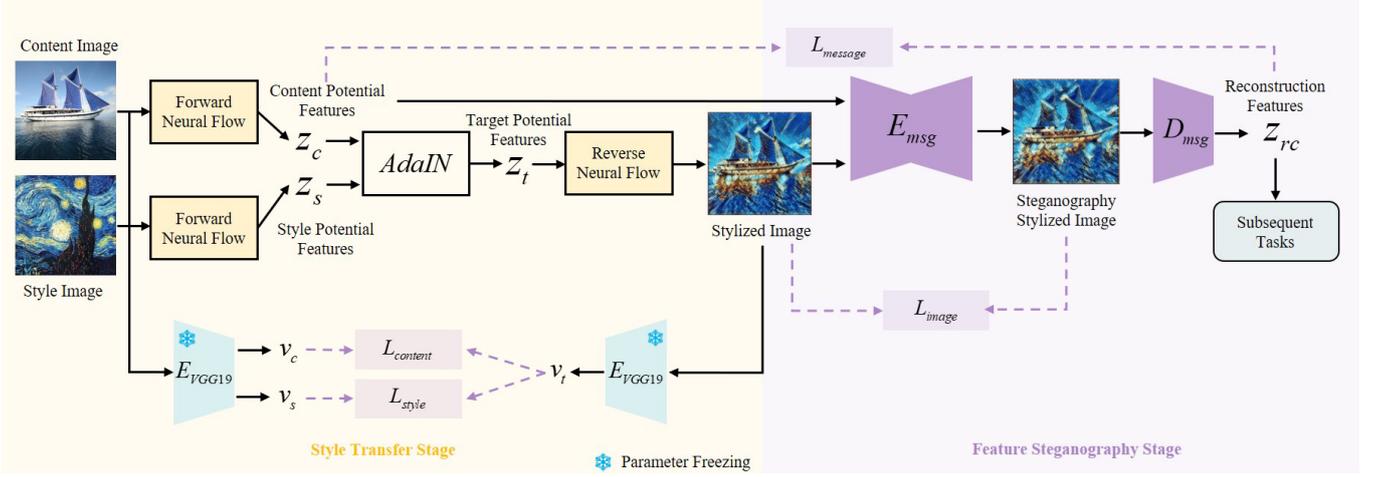

Fig. 3　The overview of proposed StyleStegan

**Proof 1** *Assuming that both $f_c$ and $f_s$ satisfy the generality of the data distribution, it follows that:*

$$f_{cs} = C(f_c)S(f_s) = \frac{f_c}{\sigma(f_c)}\sigma(f_s) \quad (3)$$

*where,*

$$C(f) = \frac{f}{\sigma(f)}, \quad S(f) = \sigma(f) \quad (4)$$

*since,*

$$\sigma(f_{cs}) = \sigma\left(\frac{f_c}{\sigma(f_c)}\right)\sigma(f_s) = 1 \times \sigma(f_s) = \sigma(f_s) \quad (5)$$

*we have,*

$$C(f_{cs}) = \frac{f_{cs}}{\sigma(f_{cs})} = \frac{f_c}{\sigma(f_c)} = C(f_c) \quad (6)$$

$$S(f_{cs}) = \sigma(f_{cs}) = \sigma(f_s) = S(f_s) \quad (7)$$

*Therefore, the adaptive instance normalization in AdaIN is an unbiased style transfer module.*

## IV. PROPOSED METHOD

### A. Overview of StyleStegan

To address the two issues mentioned, namely achieving leak-free style transfer in social networks and controlling the subsequent usage of stylized original content, we propose a feature steganography-based approach. Our method involves a two-stage model that combines style transfer and feature steganography, as depicted in Fig. 3.

During the style transfer phase, we employ a reversible neural stream-based feature network to map the features of the style image $I_s$ and the content image $I_c$ to the latent space without loss, resulting in $z_s$ and $z_c$. We then utilize the AdaIN-based unbiased style transfer module to perform stylization on the latent features. Finally, we project the target latent features back into the image using the reversible neural stream, yielding the stylized image $I_t$.

In the feature steganography phase, the steganographic network is trained to learn a feature steganography encoder. This encoder hides the content features in the latent space into the stylized image completed in the previous phase, resulting in a steganographic stylized image. Additionally, a feature extraction decoder is learned in conjunction with the feature steganography encoder. The purpose of this decoder is to extract content feature information from the steganography stylized image. The steganography stylized image is pre-embedded by the social user who owns the content copyright. Within this stage, the method offers flexibility in parameter --*stage* settings. It can be configured as a two-stage end-to-end model to complete the subsequent stylization of the social user's original content usage rights. The choice depends on the social user's preferences.

### B. Style Transfer Stage

In the style transfer stage, we borrowed the idea of the Glow model [27] to construct a reversible neural flow-based feature network. This network serves as both a deep feature extractor and an image synthesizer within the overall framework. Figure 4 illustrates the network, which comprises three reversible transformations with learning parameters: additive coupling, reversible 1 × 1 convolution, and activation normalization. It is worth noting that all components of this network are reversible, ensuring loss-free information propagation during both forward and backward propagation. The specific description of the network is provided in the following section.

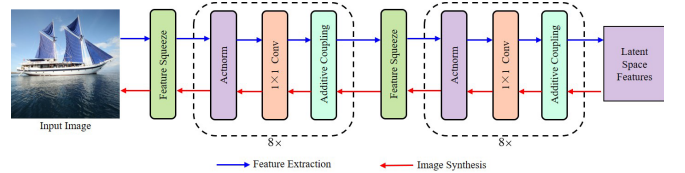

Fig. 4　Reversible neural flow based feature network

1. Additive coupling layer

[28,29] proposed an expressive and reversible layer called affine coupling. This network, however, utilizes a specific variant of affine coupling known as additive coupling. In the additive coupling layer, the parameters undergo positive calculations, as demonstrated in Equations (8) to (10):

$$x_a, x_b = split(x) \quad (8)$$
$$y_b = NN(x_a) + x_b \quad (9)$$
$$y = concat(x_a, y_b) \quad (10)$$

2. Reversible 1×1 convolutional layer

To ensure that all dimensions of the feature maps are equally affected, the channel dimensions of the feature maps in additive coupling processes should be swapped. For this purpose, a learnable reversible 1 × 1 convolutional layer is employed in the second transformation of this network, allowing for flexible channel permutation. The forward calculation of this layer is shown in Equation (11):

$$y_{i,j} = M x_{i,j} \quad (11)$$

3. Activating Normalization layer

The final transformation in this network replaces the conventional batch normalization method with activation normalization. Activation normalization involves applying an affine transformation to each channel of the tensor, as shown in Equation (12):

$$y_{i,j} = w \odot x_{i,j} + b \quad (12)$$

In this stage, the above three layers of reversible transforms combine to form a stream, and every eight streams create a block. Alongside the three transformations mentioned earlier, the compression operation divides the features into smaller feature blocks based on the spatial dimension, and then merges these blocks along the channel dimension. This compression operation is inserted between the input and block-to-block of the network, aiming to decrease the spatial size of the two-dimensional feature map. The loss function in this stage is determined by weighting the content loss and style features, which are calculated using the same equations as Equation (1) and (2), and will not be elaborated here.

C. Feature Steganography Stage

In the feature steganography stage, the network contains a steganography encoder and a corresponding steganographic decoder. The steganography encoder hides the content features $z_c$ into the stylized image $I_t$ and produces the encoded image $I_e$, which is the output of the two-stage model. And the steganography decoder tries to decode $z_c$ from $I_e$.

As a steganographic scheme, the encoded image $I_e$ and the stylized image $I_t$ should exhibit a visually imperceptible difference. To achieve this, the steganography encoder is trained to minimize the disparity between the two images. This optimization is carried out using a loss function defined as follows:

$$L_{image} = \| I_e - I_t \|_2 \quad (13)$$

The Steganography encoder consists of 1 input layer, 3 convolutional layers, and 1 output layer. It takes the stylized image as input and performs splicing embedding of steganographic features in the channel dimension. The encoder continuously optimizes the embedding of secret information and image visual quality throughout the remaining convolutional layers. Finally, it produces the steganographic stylized image as output.

The steganographic decoder is trained to optimize the extraction of content features from the steganographic stylized image. This is achieved by minimizing the loss function, as defined in Equation (14):

$$L_{message} = \| D_{msg}(I_e) - z_c \|_2 \quad (14)$$

The steganography decoder is composed of 1 input layer, 6 convolutional layers, and 1 output layer. It takes the steganographic stylized image as input and continuously optimizes the extraction accuracy of steganographic features in the 6 convolutional layers.

Finally, it outputs the extracted steganographic features. The objectives of the feature steganography phase can be summarized as Equation (15):

$$L_{steganography} = \lambda_{img} L_{image} + \lambda_{msg} L_{message} \quad (15)$$

V. EXPERIMENTS

A. Dataset and Environment

This paper follows a similar dataset setup as described in [8] to build the training set for the model. The training dataset is constructed by randomly selecting 20,000 content images from the training set of MS-COCO [30] and 20,000 style images from the training set of WikiArt [31]. To ensure consistency, the selected images were first resized to a minimum size of 512 while maintaining the aspect ratio. Subsequently, random cropping was applied to obtain images of size 256 × 256. For the test dataset, 1000 content images and 1000 style images were randomly chosen from the test sets of MS-COCO and WikiArt, respectively. During stylization, each content image was paired with one of the 10 randomly selected style images from the test set. Importantly, the test images used for experimentation were never included in the training process to prevent bias during learning and inference.

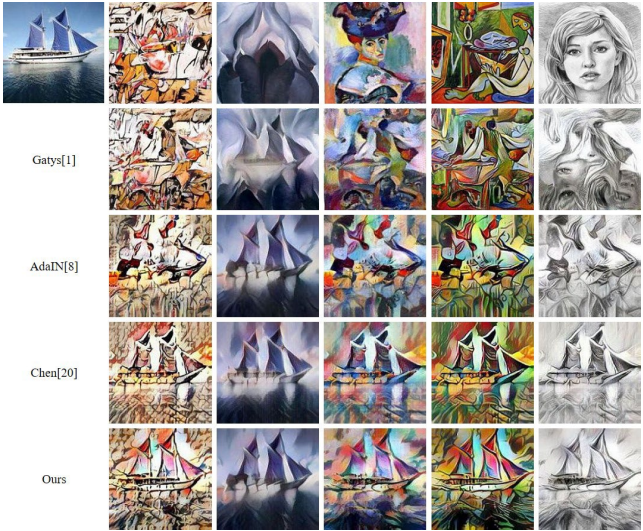

Fig. 5 Comparison of the effectiveness in serial style transfer tasks

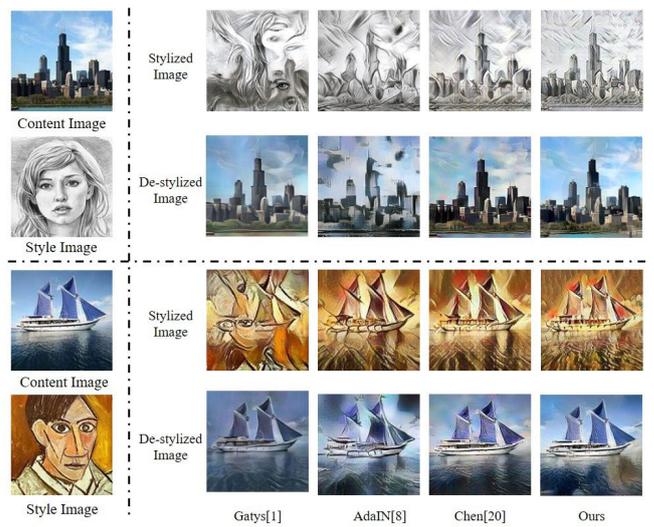

Fig. 6 Comparison of the effectiveness in reverse style transfer tasks

For the experimental environment setup, the Google online computing platform Colaboratory was used for training and inference with a Tesla T4 GPU with a video memory size of 15 GB, a system RAM size of 16 GB, a disk memory size of 128 GB, a CUDA version number of 12.0, a Python version number of 3.8, and the deep learning framework used was Pytorch, whose version is 2.0.0+cu118.

### B. Visual Comparison

In the experimental design, we compare the proposed model in this paper with the baseline model of [1], the adaptive instance normalized baseline model of [8], and the end-to-end model of [30]. These comparisons are made to validate the qualitative results through serial and reverse style transfer tasks.

1. Serial style transfer

Serial style transfer aims to apply a new style to a stylized image while minimizing the impact of previous stylizations. The desired outcome of serial style transfer is to achieve an artistic rendering effect similar to directly and separately stylizing the original content with the new style. In the experiments conducted for this task, we compare the results of our method with existing methods, as depicted in Figure 7.

The obtained results demonstrate that our method effectively reduces the corruption of content features caused by intermediate style transfers. This is achieved by selectively and discreetly embedding content features into each stylized image,
catering to the specific requirements of users for subsequent style transfers. By doing so, the original content features can be extracted and utilized in each style transfer operation.

2. Reverse style transfer

The goal of reverse style transfer is to de-stylize a stylized image while preserving its original content as much as possible. Ideally, the outcome of reverse style transfer should closely resemble the input content image. However, the baseline model of Gatys et al [1] and AdaIN [8] do not provide a direct approach for reverse style transfer using the original content image as the reference for the target style and applying the stylization to the given stylized image. This method requires access to the original content image. In contrast, the model developed by Chen et al [20] and the present method can extract the hidden features necessary for de-styling using only the given stylized image. To evaluate the effectiveness of the present method in reverse style transfer, it is compared with existing methods, and the experimental results are presented in Figure 8. The results demonstrate that the proposed method achieves better content image reconstruction in the reverse style transfer task.

### C. Quantitative Comparison

Quantitative experiments entail the application of different models for conducting serial style transfer and reverse style transfer tasks. These models generate output images, which are subsequently compared to their respective desired images. The aim of this comparison is to assess the performance of the models in both serial style transfer and reverse style transfer. The findings from these quantitative experiments are presented in Table 1.

Tab. 1 Quantitative experimental Results of serial and reverse style transfer

|         |        | Gatys[1] | AdaIN[8] | Chen[20] | Ours |
|---------|--------|----------|----------|----------|------|
| Serial  | L2 ↓   | 7.5239   | 0.0213   | 0.0148   | **0.0104** |
|         | SSIM ↑ | 0.0473   | 0.5471   | 0.7143   | **0.8213** |
|         | LPIPS ↓| 0.4316   | 0.4614   | 0.2437   | **0.1483** |
| Reverse | L2 ↓   | 4.4331   | 0.0368   | **0.0187** | 0.0193 |
|         | SSIM ↑ | 0.2041   | 0.3826   | 0.4796   | **0.5145** |
|         | LPIPS ↓| 0.3694   | 0.4628   | **0.3323** | 0.3802 |

## VI. CONCLUSIONS

We address the issue of content leakage that commonly arises in serial and reverse style transfer and propose a corresponding solution. Our experimental results demonstrate the superiority of our method in both serial and reverse style transfer tasks. Specifically, our method achieves a 14.98% and 7.28% higher SSIM compared to the suboptimal performing baseline model for the two tasks, respectively. The specific contributions of our work are as follows:

1. We analyze the phenomenon of content leakage in typical style transfer methods and identify image reconstruction loss, decoding training bias, and transfer module bias as potential causes.

2. To ensure content integrity and style consistency in stylized images, we design loss-free and bias-free reversible neural flow networks. These networks eliminate content reconstruction loss during encoding and decoding, as well as feature bias error during style transfer.

3. We develop a model for content feature steganography and extraction, which allows the embedding of content features into stylized images. This enables subsequent high-quality stylization while ensuring that only authorized users with specific extraction methods can retrieve the original content, providing control over the stylized images.

4. We conduct experimental validation on the MSCOCO and Wikiart public datasets to showcase the visual effectiveness and outstanding performance of our method in generating stylized images. Our results provide solid evidence for the effectiveness of our approach in addressing the content leakage problem in serial and reverse style transfer tasks.

## ACKNOWLEDGMENT

Xiujian Liang and Binghuang Liu acknowledge the support from the National Student Innovation and Entrepreneurship Training Program Support Project of China (202210300028Z). Thanks to the Multimedia Artificial Intelligent Security Laboratory of Fudan University for the research direction guidance and experimental support.